\documentclass[journal]{IEEEtran}
\usepackage{amsmath,amsfonts}
\usepackage{algorithmic}
\usepackage{array}
\usepackage[caption=false,font=normalsize,labelfont=sf,textfont=sf]{subfig}
\usepackage{textcomp}
\usepackage{stfloats}
\usepackage{url}
\usepackage{verbatim}
\usepackage{graphicx}
\usepackage{tikz}
\usepackage{comment}
\usepackage{amssymb} % define this before the line numbering.
\usepackage{color}
\usepackage{colortbl}
\usepackage{booktabs}
\usepackage{multirow}
\usepackage{wrapfig}
\usepackage{pifont}% http://ctan.org/pkg/pifont
\usepackage{ragged2e}
\usepackage{url}
\usepackage{xcolor}
\usepackage{orcidlink}
\usepackage{marvosym}
\usepackage{algorithm}
\usepackage{algorithmic}
\definecolor{tabfirst}{rgb}{1, 0.7, 0.7}
\definecolor{tabsecond}{rgb}{1, 0.85, 0.7}
\definecolor{tabthird}{rgb}{1, 1, 0.7}
\usepackage{hyperref}
\hypersetup{
    colorlinks=true,
    citecolor=blue,
    urlcolor=blue,
    linkcolor=blue
}
\usepackage[]{hyperref}

\def\BibTeX{{\rm B\kern-.05em{\sc i\kern-.025em b}\kern-.08em
    T\kern-.1667em\lower.7ex\hbox{E}\kern-.125emX}}
\usepackage{balance}

\usepackage{amsmath,amsfonts,bm}

\def\eqref#1{equation~\ref{#1}}
\def\1{\bm{1}}

\DeclareMathAlphabet{\mathsfit}{\encodingdefault}{\sfdefault}{m}{sl}
\SetMathAlphabet{\mathsfit}{bold}{\encodingdefault}{\sfdefault}{bx}{n}

\usepackage{xcolor}
\usepackage{wrapfig}
\usepackage{pifont}
\usepackage{xspace}

\newcommand{\framework}{\textbf{DreamVideo-Omni}\xspace}
\newcommand{\frameworkplain}{DreamVideo-Omni\xspace}

\newcommand{\tabincell}[2]{\begin{tabular}{@{}#1@{}}#2\end{tabular}}

\newlength\savewidth

\newcommand{\eg}{\textit{e.g.}}
\newcommand{\ie}{\textit{i.e.}}

\definecolor{lightgreen}{rgb}{0.56, 0.93, 0.56} 
\definecolor{mycolor_blue}{HTML}{E7EFFA}
\definecolor{mycolor_green}{HTML}{E6F8E0}
\definecolor{mycolor_gray}{HTML}{ECECEC}
\definecolor{pearDark}{HTML}{2980B9}

\definecolor{darkred}{rgb}{0.7,0.1,0.1}
\definecolor{darkgreen}{rgb}{0.1,0.6,0.1}
\newcommand{\cmark}{\color{darkgreen}{\ding{51}}}
\newcommand{\xmark}{\color{darkred}{\ding{55}}}

\begin{document}
\title{\frameworkplain: Omni-Motion Controlled Multi-Subject Video Customization with Latent Identity Reinforcement Learning}
\author{
Yujie Wei$^*$\orcidlink{0009-0003-9304-0609}, Xinyu Liu$^*$\orcidlink{0009-0007-0456-921X}, Shiwei Zhang$^\dagger$\orcidlink{0000-0002-6929-5295}, Hangjie Yuan\orcidlink{0009-0009-3270-1526}, Jinbo Xing\orcidlink{0000-0002-2181-1879}, Zhekai Chen\orcidlink{0009-0005-1369-5483}, Xiang Wang\orcidlink{0000-0003-0785-3367}, Haonan Qiu\orcidlink{0000-0002-3878-1418}, Rui Zhao\orcidlink{0000-0003-4271-0206}, Yutong Feng\orcidlink{0000-0003-0575-6790}, Ruihang Chu\orcidlink{0000-0001-9057-745X}, Yingya Zhang, Yike Guo\orcidlink{0009-0005-8401-282X},~\IEEEmembership{Fellow,~IEEE},\\
Xihui Liu\orcidlink{0000-0003-1831-9952},~\IEEEmembership{Member,~IEEE} and Hongming Shan$\textsuperscript{\Letter}$\orcidlink{0000-0002-0604-3197},~\IEEEmembership{Senior Member,~IEEE}
\thanks{
$^*$Equal Contribution\quad$^\dagger$Project Leader\quad$\textsuperscript{\Letter}$Corresponding Author}
\thanks{Email: Yujie Wei \texttt{yjwei22@m.fudan.edu.cn} and Xinyu Liu \texttt{xliugd@connect.ust.hk}}
\thanks{Yujie Wei and Hongming Shan are with Fudan University.
Xinyu Liu and Yike Guo are with The Hong Kong University of Science and Technology.
Shiwei Zhang, Jinbo Xing, Xiang Wang, Yutong Feng, Ruihang Chu, Yingya Zhang are with Tongyi Lab, Alibaba Group.
Hangjie Yuan is with Zhejiang University.
Zhekai Chen and Xihui Liu are with MMLab, The University of Hong Kong.
Haonan Qiu is with Nanyang Technological University.
Rui Zhao is with Show Lab, National University of Singapore.}
}

\maketitle

\begin{abstract}
While large-scale diffusion models have revolutionized video synthesis, achieving precise control over both multi-subject identity and multi-granularity motion remains a significant challenge.
Recent attempts to bridge this gap often suffer from limited motion granularity, control ambiguity, and identity degradation, leading to suboptimal performance on identity preservation and motion control.
In this work, we present \framework, a unified framework enabling harmonious multi-subject customization with omni-motion control via a progressive two-stage training paradigm.
In the first stage, we integrate comprehensive control signals for joint training, encompassing subject appearances, global motion, local dynamics, and camera movements.
To ensure robust and precise controllability, we introduce a condition-aware 3D rotary positional embedding to coordinate heterogeneous inputs and a hierarchical motion injection strategy to enhance global motion guidance.
Furthermore, to resolve multi-subject ambiguity, we introduce group and role embeddings to explicitly anchor motion signals to specific identities, effectively disentangling complex scenes into independent controllable instances.
In the second stage, to mitigate identity degradation, we design a latent identity reward feedback learning paradigm by training a latent identity reward model upon a pre-trained video diffusion backbone.
This provides motion-aware identity rewards in the latent space, prioritizing identity preservation aligned with human preferences.
Supported by our curated large-scale dataset and the comprehensive \textbf{DreamOmni Bench} for multi-subject and omni-motion control evaluation, \frameworkplain demonstrates superior performance in generating high-quality videos with precise controllability.
Our project webpage: \url{https://dreamvideo-omni.github.io}.
\end{abstract}


\begin{figure*}[t]
  \centering
  \includegraphics[width=1.0\linewidth]{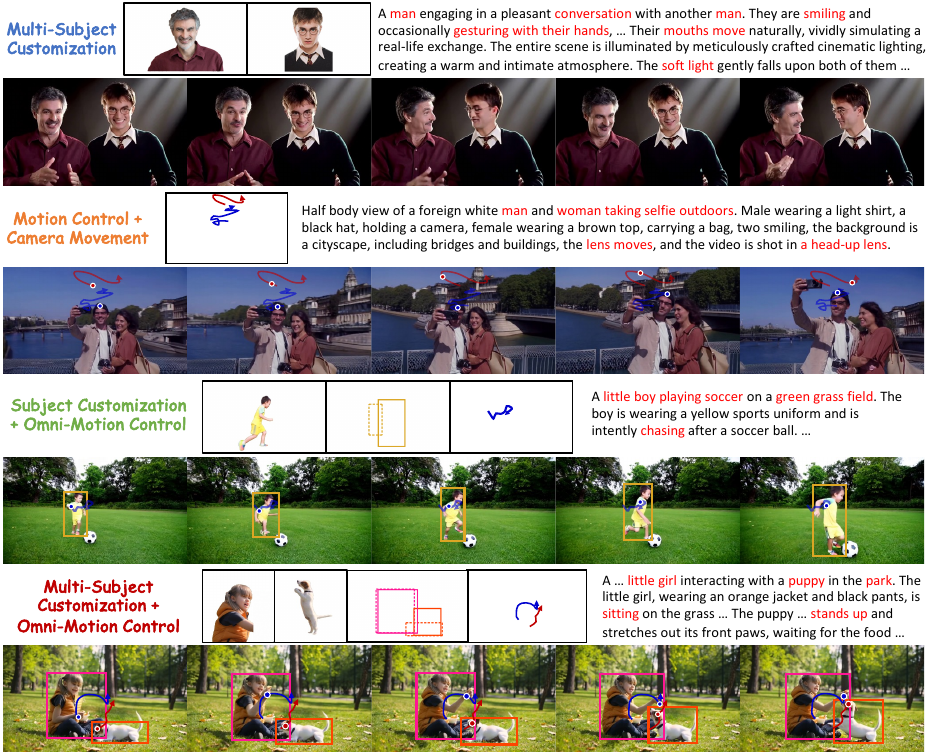}
  \caption{\textbf{Zero-shot multi-subject customization and omni-motion control achieved by \framework.} Our method enables seamless multi-subject customization, precise motion and camera control, and simultaneous single/multi-subject customization with omni-motion control.}
  \label{fig:teaser}
\end{figure*}

\section{Introduction}
\label{sec:intro}

The landscape of video generation has been revolutionized by the advent of diffusion models~\cite{VDM, LVDM, modelScope, videocrafter1, wang2023lavie, hong2022cogvideo, yang2024cogvideox, wan2025wan, wei2025routing}.
While these foundation models demonstrate impressive capability in synthesizing high-fidelity videos from textual descriptions, the demand for real-world applications necessitates more granular control. Specifically, users often require the generation of videos that simultaneously preserve the identities of multiple subjects while adhering to comprehensive motion control, including global object motion, local limb motion, and camera movements.

Despite the remarkable advances in controllable video generation, achieving this dual objective---robust multi-subject ID preservation and precise, multi-granularity motion control---remains an open challenge.
Existing approaches typically diverge into two independent research directions.
On one hand, subject-driven methods~\cite{jiang2024videobooth, wei2024dreamvideo, liu2025phantom} utilize adapters or tuning-free mechanisms to inject appearance, yet they often lack the capacity for precise spatial control, resulting in videos where subjects drift uncontrollably or remain static.
On the other hand, motion-controlled methods~\cite{wang2024motionctrl, wang2024boximator, xing2025motioncanvas, chu2025wan} excel at guiding movement via bounding boxes or trajectories but fail to achieve omni-motion control and customize user-specified subjects, limiting their practical applicability.

Recent works attempt to integrate these capabilities within a unified framework, aiming to synthesize videos that faithfully preserve subject identities while adhering to specified motion patterns~\cite{wei2024dreamvideo2, ju2025fulldit, cai2025omnivcus, zhang2025tora2}.
However, these methods often yield suboptimal performance due to the intrinsic trade-off between subject preservation and motion control. This limitation manifests primarily in three aspects:
\textbf{1) Limited Motion Control Granularity.} 
Most existing methods rely on a single type of motion signal, such as bounding boxes~\cite{wei2024dreamvideo2}, depth maps~\cite{ju2025fulldit, cai2025omnivcus}, or sparse trajectories~\cite{zhang2025tora2}, to guide generation. This restricted conditioning fails to support the simultaneous control of global object placement, fine-grained local dynamics, and camera movement, thereby limiting the flexibility and realism of the generated videos.
\textbf{2) Ambiguity in Motion Control.} 
Current approaches typically inject all conditioning signals indiscriminately without explicit binding mechanisms~\cite{ju2025fulldit, cai2025omnivcus}. In multi-subject scenarios, this leads to severe ambiguity, as the model struggles to discern which motion pattern corresponds to which specific reference subject. This confusion is further exacerbated when integrating multi-granular motion controls.
\textbf{3) Identity Degradation.} 
Compared to independent subject customization, introducing motion control often compromises identity fidelity. This stems from the divergent nature of the objectives: identity preservation encourages pixel-level consistency with a \textit{static} reference image, whereas motion control necessitates \textit{dynamic} pixel variation and temporal evolution to render movement. Standard diffusion reconstruction losses are insufficient to reconcile this conflict, leading to degradation of fine-grained identity details, particularly when synthesizing large-amplitude motions.

To address these issues, we posit that it is pivotal to simultaneously enhance motion controllability and identity preservation. 
First, motion control signals must be explicitly bound to their corresponding reference subjects to resolve ambiguity, thereby facilitating precise, multi-granular control.
Second, to further reinforce identity fidelity, the learning objective should be aligned with human preferences. 
We recognize that subject customization is inherently subjective and distinct from rigid pixel-wise correspondence, as a subject's visual appearance naturally varies with viewpoints and poses, yet their identity remains consistent.
Consequently, the optimization process should prioritize perceptual alignment with human experience rather than relying solely on low-level reconstruction metrics.

Based on these insights, we propose \framework, a unified framework enabling harmonious multi-subject customization and omni-motion control through a progressive two-stage training paradigm illustrated in Fig.~\ref{fig:teaser}.
In the first stage, referred to as  \textit{omni-motion and identity supervised fine-tuning}, we integrate comprehensive control signals, formulated as structured triplets of $\langle$\textit{Reference Subject, Global Box, Local Trajectory}$\rangle$, into a single DiT architecture.
To effectively process these heterogeneous inputs, we design a \textit{condition-aware 3D Rotary Positional Embedding (RoPE)} that assigns distinct spatiotemporal indices to diverse conditions, facilitating faster convergence and enhanced training stability.
To guarantee precise global motion control, we employ a \textit{hierarchical motion injection} strategy, infusing bounding box conditions into each transformer block to reinforce spatial guidance.
Furthermore, to resolve control ambiguity in multi-subject scenarios, we introduce learnable \textit{group and role embeddings} that distinguish distinct control units and specify signal modalities, explicitly anchoring motion signals to their corresponding identities.
Collectively, these architectural designs establish a foundation for integrating subject customization with controllable motion generation within a unified framework.

In the second stage, referred to as \textit{latent identity reward feedback learning}, we move beyond standard diffusion losses and employ a reward feedback strategy within the latent space to mitigate identity degradation during dynamic motion generation.
Specifically, we train a specialized Latent Identity Reward Model (LIRM) to provide rewards.
Departing from previous methods that rely on static image encoders (\eg, CLIP~\cite{clip} or DINO~\cite{dino}) as reward models, which overlook temporal dynamics, our LIRM is constructed upon a pre-trained Video Diffusion Model (VDM), yielding two advantages:
1) Motion-Aware Identity Assessment: By leveraging the VDM's inherent spatiotemporal priors, the reward model evaluates video-level identity consistency that integrates motion dynamics, penalizing static ``copy-paste'' artifacts while encouraging robust identity preservation under large motion.
2) Computationally Efficient Training: Computing rewards in the latent space bypasses expensive VAE decoding. This enables direct gradient backpropagation from reward models to the video generation model, fully leveraging the potential of reward feedback learning.

Remarkably, this progressive two-stage training paradigm enables the seamless composition of multiple tasks while naturally facilitating the emergence of novel generative abilities. Despite being built upon a text-to-video base model, our framework spontaneously unlocks zero-shot image-to-video generation and first-frame-conditioned trajectory control.

Finally, to support the training of our unified framework, we curate a large-scale dataset for multi-subject customization and omni-motion control, which comprises 2M video clips, enriched with multi-subject reference images, bounding boxes, and trajectory conditions.
Beyond training resources, we also address the critical lack of comprehensive evaluation protocols. Existing benchmarks either isolate customization from controllable generation~\cite{wei2024dreamvideo, jiang2024videobooth} or focus exclusively on simple point trajectories~\cite{chu2025wan}. To bridge this gap, we construct the \textbf{DreamOmni Bench}, a holistic evaluation suite composed of 1,027 high-quality, real-world videos. This benchmark explicitly categorizes single- and multi-subject scenarios and is equipped with dense annotations, enabling the first unified evaluation of identity preservation and complex motion controllability in zero-shot settings.

In summary, our contributions are five-fold:
\textit{1)} We present \framework, the first unified framework that harmoniously integrates multi-subject customization with omni-motion control within a single DiT architecture.
\textit{2)} We propose specialized architectural components to ensure precise controllability. 
Specifically, we introduce group and role embeddings to resolve multi-subject ambiguity, condition-aware 3D RoPE to coordinate heterogeneous inputs for stable training, and hierarchical motion injection to enhance global motion control.
\textit{3)} We design a latent identity reward feedback learning paradigm. We train a VDM-based latent identity reward model to evaluate motion-aware identity preservation, effectively mitigating identity degradation under large motion.
\textit{4)} We establish \textbf{DreamOmni Bench}, a new benchmark consisting of over 1K curated videos with comprehensive annotations, designed to simultaneously quantify multi-subject consistency and motion control precision. 
We also design a comprehensive data processing pipeline for multi-subject customization and omni-motion control tasks.
\textit{5)} Extensive experimental results demonstrate our framework's superiority over state-of-the-art methods in both identity preservation and motion control.

\section{Related Work}
\label{sec:related_work}

\noindent\textbf{Customized video generation.}\quad
Customized image generation has garnered growing attention~\cite{disenbooth, wei2023elite, shi2024instantbooth, ruiz2024hyperdreambooth, hua2023dreamtuner, han2024face_adapter, mix_of_show, fastcomposer, customDiffusion, zhou2024storydiffusion, li2024photomaker, huang2024realcustom}. Recently, many works explore customized video generation using a few subject or facial images~\cite{dreamix, chefer2024still_moving, ma2024magicme, he2024id_animator, zhou2024sugar, wu2024videomaker, huang2024dive, she2025customvideox, wu2024customcrafter, zhang2025fantasyid,yuan2025identity, wei2025dreamrelation}, while several works study the challenging multi-subject video customization task~\cite{wang2024customvideo, chen2024disenstudio, chen2025multi, huang2025conceptmaster}.
For example,
ConsisID~\cite{yuan2025identity} leverages frequency decomposition to decouple facial contours and details in video DiT for consistent identity across frames.
For multi-subject customization, VideoMage~\cite{huang2025videomage} and Video Alchemist~\cite{chen2025multi} extend single-subject methods to open-set personalization, improving multi-subject identity consistency without test-time tuning. 
However, these methods focus on independent subject customization and often struggle with ``copy-paste'' artifacts and limited controllability, restricting their applicability in real-world scenarios.
Considering that spatial content and temporal dynamics are two essential components of videos, DreamVideo~\cite{wei2024dreamvideo} customizes both subject and motion by training two adapters and combining them at inference, while 
MotionBooth~\cite{wu2024motionbooth} fine-tunes a model to learn subjects and edits attention maps to control motion during inference.
More recently, Tora2~\cite{zhang2025tora2} integrates trajectory control into subject customization by introducing a decoupled personalization extractor and a gated self-attention mechanism.
However, these methods rely predominantly on standard diffusion losses and suffer from the trade-off between motion control and identity preservation, often resulting in identity degradation under large-amplitude motions.
In contrast, \frameworkplain effectively resolves this dilemma by aligning identity preservation with human preferences. By introducing a latent identity reward feedback learning paradigm and training a specialized latent reward model, our approach ensures harmonious multi-subject customization with omni-motion control.

\noindent\textbf{Motion control in video generation.}\quad
Recent advancements in video generation focus on enhancing motion dynamics using extra control signals~\cite{shi2024motion, wu2025draganything, bahmani2025tc4d, namekata2024sg, geng2025motion, zheng2024cami2v, hu2024comd, yesiltepe2024motionshop, niu2025mofa, jeong2024dreammotion, li2024motrans, feng2024i2vcontrol, park2024spectral}.
Many motion customization methods learn motions from reference videos~\cite{jeong2024vmc, customize_a_video, DMT, motionInversion, lamp}, 
but require complicated fine-tuning during inference.
To alleviate this, some training-free approaches~\cite{jain2024peekaboo, direct_a_video, ma2023trailblazer, motion_zero, huang2025zero} employ attention manipulation or guidance to achieve zero-shot control.
However, these methods often sacrifice motion precision and temporal consistency in complex scenarios. 
In contrast, several works use trajectories or coordinates as additional conditions to train a motion control module ~\cite{yin2023dragnuwa, wang2024motionctrl, wang2024boximator, imageConductor, chu2025wan,zhang2025motionpro, zhang2025tora, xing2025motioncanvas}.
For example,
Motion Prompting~\cite{geng2025motion} conditions generation on spatio-temporal trajectories for camera control and motion transfer, with prompt expansion for complex inputs. MagicMotion~\cite{li2025magicmotion} uses object masks and bounding boxes to control motion, trained on a pretrained image-to-video diffusion model.
Wan-Move~\cite{chu2025wan} uses dense point trajectories projected into latent space to propagate features, enabling motion control based on the first frame in image-to-video models.
Nonetheless, they are incapable of delivering comprehensive fine-grained motion control, such as the simultaneous control of global motion, local dynamics, and camera movements. Moreover, they fail to incorporate user-specified subject appearances, which restricts their real-world applicability.
In contrast, our \frameworkplain unifies user-specified subject customization and multi-granular motion control into a single framework, incorporating a binding mechanism to explicitly resolve motion ambiguity and ensure precise control.

\noindent\textbf{Identity-based reinforcement learning.}\quad
To ensure identity consistency in customized video generation, recent research has increasingly integrated reinforcement learning into optimization frameworks~\cite{li2025magicid,shen2025identity,meng2025identity,zhu2025aligning}.
For instance, MagicID~\cite{li2025magicid} employs DPO~\cite{rafailov2023direct} to enhance text-to-video identity stability, though it still necessitates costly per-identity LoRA adaptation and test-time fine-tuning.
More recently, Identity-GRPO~\cite{meng2025identity} leverages human preference-driven GRPO~\cite{shao2024deepseekmath} to preserve stable facial features during complex interactions by constructing multi-character reward models.
IPRO~\cite{shen2025identity} adopts the Reward Feedback Learning (ReFL)~\cite{xu2023imagereward} paradigm. It backpropagates gradients from similarity-based rewards directly into the diffusion model, bypassing explicit reward model training or identity-specific tuning.
However, both Identity-GRPO and IPRO require decoding latents into pixel space for reward calculation, which incurs heavy GPU overhead and restricts feedback to the final denoising steps, resulting in limited performance improvements.
In contrast, we conduct identity-driven reinforcement learning directly within the latent space, significantly reducing computational overhead.
While the very recent general video generation method PRFL~\cite{mi2025video} also introduces latent-space reward modeling to mitigate computational costs, it primarily focuses on optimizing general motion quality, lacking the capacity to distinguish and preserve intricate subject identities.
Unlike PRFL, we process the generated video and the reference image in parallel under varying noise levels, explicitly leveraging the clean reference latents as queries to attend to the noisy video latents for computing identity rewards.
This design substantially enhances identity customization capabilities by incorporating identity information across the full range of timesteps, thereby fully unleashing the potential of ReFL in customization tasks.

\begin{figure*}[t]
  \centering
  \includegraphics[width=0.9\linewidth]{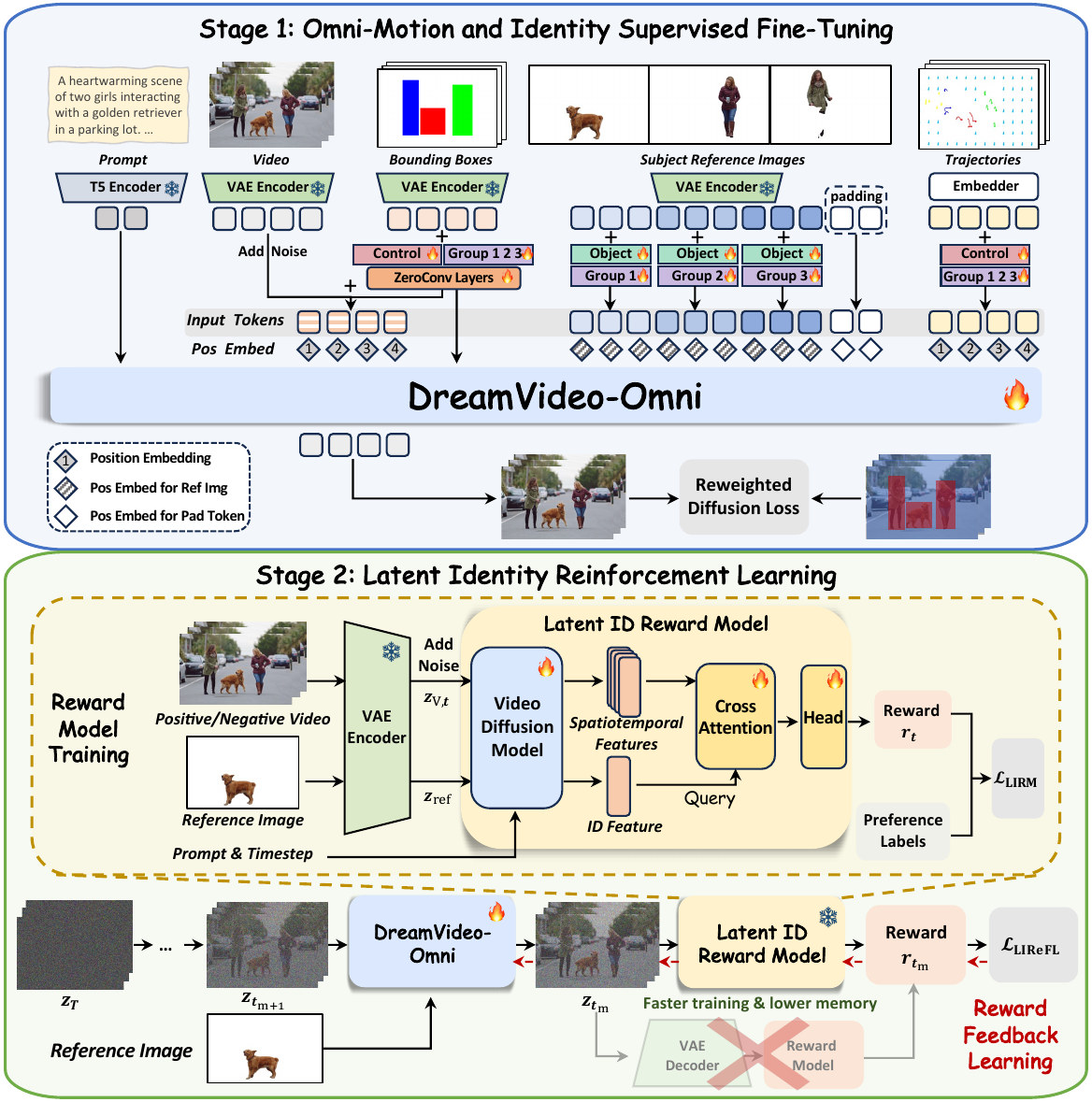}
  \caption{\textbf{Overview of \frameworkplain.} 
  In Stage 1, the framework introduces an all-in-one video DiT that incorporates reference images, bboxes, and trajectories for multi-subject customization and omni-motion control. Stage 2 further enhances identity fidelity via the proposed latent identity reward feedback learning mechanism, which utilizes a latent identity reward model to directly evaluate intermediate latents, completely bypassing the expensive VAE decoder for faster training.
  }
  \label{fig:framework}
\end{figure*}

\section{Our mehtod: \frameworkplain}
\label{sec:method}

Given a set of reference images and motion signals (\ie, bounding boxes and trajectories), our \frameworkplain employs a unified video diffusion transformer to jointly condition on multi-subject appearances, global and local object motions, and camera movements, enabling flexible compositional video generation without test-time fine-tuning. The overall pipeline is shown in Fig.~\ref{fig:framework}, which follows a progressive two-stage training paradigm. In the first omni-motion and identity supervised fine-tuning stage, we train an all-in-one video DiT to integrate heterogeneous control signals via hierarchical injection, condition-aware 3D RoPE, and group/role embeddings, effectively resolving multi-subject ambiguity and enabling fine-grained motion control. In the second latent identity reinforcement learning stage, we employ latent identity reward feedback learning, where a specialized latent identity reward model provides direct identity supervision on intermediate noisy latents, efficiently bypassing the computationally expensive VAE decoding. In the following sections, we detail this two-stage paradigm in Secs.~\ref{sec:method_sft} and~\ref{sec:method_rl}. Subsequently, we describe the dataset construction in Sec.~\ref{sec:data_construction} and introduce our newly constructed DreamOmni Bench in Sec.~\ref{sec:benchmark}.

\subsection{Omni-Motion and Identity Supervised Fine-Tuning}
\label{sec:method_sft}
This subsection will dive into the key designs of \frameworkplain, including model architecture, task design, and details of the designed components.

\subsubsection{Model Architecture and Task Design}
As illustrated in Fig.~\ref{fig:framework}, we instantiate an all-in-one framework by adapting a pre-trained text-to-video DiT~\cite{wan2025wan}.
Our model is jointly trained on a comprehensive set of tasks, including single- and multi-subject customization, global and local object motion control, and camera movement control.
To enable precise and flexible composition of these tasks, we carefully craft four compact, interaction-friendly conditioning signals.
\begin{enumerate}
\item[1)] \textbf{Subject appearance.} We use one reference image per subject and segment it to obtain the image with a blank background, preserving distinct identity features while reducing background
interference.
\item[2)] \textbf{Global object motion.} We employ scene-anchored bounding boxes to indicate global object motion. These boxes serve as an intuitive proxy for scene-level spatial attributes, effectively capturing object position, scale, aspect ratio, and relative depth.
In practice, users can simply specify start and end boxes with optional intermediate boxes to achieve flexible global motion control.
\item[3)] \textbf{Local object motion.} We represent local object motion using sparse point-wise trajectories. Compared to global motion, local object motion targets finer-grained, complex in-place dynamics (\eg, limb raises, head turns) that enrich per-object motion details. This point-based control flexibly captures complex non-rigid deformations.
\item[4)] \textbf{Camera movement.} We also use point-wise trajectories to achieve camera movement control.
Prior works typically rely on explicit 3D camera parameters and auxiliary datasets for training~\cite{wang2024motionctrl, cai2025omnivcus}. 
While effective, these approaches increase training cost and hinder usability.
In contrast, we observe that camera movement can be effectively induced by applying point-wise trajectories to background pixels.
This allows us to unify camera and local motion control under the same trajectory conditioning mechanism, reducing training overhead while improving interactivity.
\end{enumerate}

\subsubsection{Conditioning Signal Injection}

Beyond the task formulation, the effective injection of conditioning signals is pivotal for achieving precise control.

\noindent\textbf{1) Subject appearance.}\quad  To enable robust multi-subject customization while mitigating ``copy-paste'' artifacts, reference images are extracted from frames temporally disjoint from the current training clip and undergo extensive data augmentation, detailed in Sec.~\ref{sec:arch_components}.
These processed images are then encoded via a 3D VAE, and their latent representations are concatenated to form the conditioning input for the video DiT.

\noindent\textbf{2) Global object motion.}\quad
Before inputting bounding box sequences into the DiT, we first filter out clips exhibiting abrupt object fluctuations by detecting adjacent-frame bounding box IoU to ensure training stability.
The valid sequences are then rendered as RGB videos on a white background, where each object is assigned a unique random color and pixel-wise averaging is applied in overlapping regions to resolve ambiguity.
These rendered videos are projected into the latent space via a 3D VAE.
To enable efficient and effective control signal injection, we implement a \textit{hierarchical motion injection strategy}: bounding box latents are added to both the noisy input latents and the output of each DiT block via learnable, layer-specific zero-convolutions, formulated as:
\begin{align}
\bm{h}_0 = \bm{z}_t + \mathcal{Z}_{\text{in}}(\bm{z}_{\text{box}}), \quad
\bm{h}_{l+1} = \text{Block}_l(\bm{h}_{l}) + \mathcal{Z}_{l}(\bm{z}_{\text{box}}),
\end{align}
where $\bm{z}_t$ and $\bm{z}_{\text{box}}$ are the input noisy video latents and bounding box latents, respectively. $\mathcal{Z}_{\text{in}}$ and $\mathcal{Z}_{l}$ denote the zero-convolutions at the input stage and the $l$-th DiT block. $\bm{h}_{l}$ is the input hidden state to the $l$-th block.
This dense injection mechanism enhances the precision of global motion control without increasing the token sequence length.

\noindent\textbf{3) Local object motion and camera movement.}\quad  Since we employ point trajectories to control both local object dynamics and camera movements, the sampled points must comprehensively cover both foreground subjects and background regions.
While uniform sampling is an intuitive solution, it often struggles with the fine-grained motion control due to insufficient sampling density on object boundaries.
To address this, we devise a hybrid sampling strategy that stochastically alternates between two modes: (i) random grid sampling, which ensures broad coverage of whole scene dynamics (background and objects); and (ii) object-aware sampling, which samples strictly within foreground masks to focus on intricate local dynamics.
To improve robustness on trajectory densities, a subset of trajectories is randomly dropped during training.
Following Motion Prompting~\cite{geng2025motion}, we construct trajectory tokens by generating unique sinusoidal positional encodings and scattering them into blank feature maps according to their discretized spatiotemporal coordinates. These tokens are subsequently concatenated with the noisy video and reference image latents to condition the DiT training.

\subsubsection{Specialized Architectural Components }
\label{sec:arch_components}
\noindent\textbf{Condition-aware 3D RoPE.}\quad 
To process heterogeneous inputs, including video latents, multi-subject reference images, and motion control signals, within a unified DiT architecture, we concatenate all latent tokens along the temporal dimension and design a condition-aware 3D Rotary Positional Embedding (RoPE).
While our condition-aware 3D RoPE maintains standard indexing for spatial dimensions to preserve geometric structure, it employs a specialized temporal indexing strategy to distinguish input types:

\textit{(i) Video frame tokens:} We assign sequential temporal indices $t \in [0, T-1]$ to these tokens to ensure temporal consistency. Note that bounding box latents are element-wise added to these frames, thereby naturally inheriting the same positional embeddings.

\textit{(ii) Reference image tokens:} We assign a shared, distinct time index $t_{\text{ref}}$ to all valid reference image tokens. This design explicitly decouples reference subjects from the video tokens, 
instructing the model to treat these tokens as static visual conditions rather than sequential frames.

\textit{(iii) Padding tokens:} To handle varying numbers of reference subjects across videos, we pad the reference image tokens to a fixed capacity $N_{\text{max}}$ within the batch. These padded tokens are assigned a distinct ``invalid'' time index $t_{\text{pad}}$, allowing the model to identify and ignore these non-informative tokens.

\textit{(iv) Trajectory tokens:} To provide precise pixel-level motion control, trajectory tokens inherit the same temporal indices $t \in [0, T-1]$ as the video frame tokens, ensuring strict spatiotemporal alignment with the corresponding video.

By integrating these specific indices into the 3D RoPE, we enable the unified DiT architecture to effectively process heterogeneous inputs, including both reference images and diverse motion control signals.

\noindent\textbf{Group and role embeddings.}\quad 
To mitigate control ambiguity in multi-subject generation and distinguish the functional roles of heterogeneous inputs, we introduce two types of learnable embeddings: group embeddings and role embeddings.
\textit{First}, we formulate the fundamental control unit as a triplet of $\langle\text{Reference Subject}, \text{Global Box}, \text{Local Trajectory}\rangle$, and assign a unique \textit{group embedding} to each unit.
Given that a reference image captures a single subject, its corresponding group embedding is added to all latent tokens of the image.
In contrast, for the bounding box and trajectory latents, this same group embedding is injected exclusively into the spatial regions and track points corresponding to the subject.
This explicit binding mechanism ensures that each subject is correctly associated with its corresponding bounding box and trajectories, effectively preventing control confusion.
\textit{Second}, we introduce role embeddings, comprising object and control embeddings, to differentiate input signals' functionalities.
Specifically, an \textit{object embedding} is added to all reference image tokens to designate them as visual appearance assets, whereas a \textit{control embedding} is applied to all bounding box and trajectory tokens to mark them as motion control guidance. This functional distinction enables the model to effectively process heterogeneous conditions.

\noindent\textbf{Data augmentation for subject customization.}\quad  To mitigate ``copy-paste'' artifacts caused by directly training on reference subjects, we apply a robust augmentation pipeline to the reference images. Specifically, we stochastically employ geometric transformations (\eg, flipping, rotation, affine shearing, and cropping) and visual degradations (\eg, blur, color jitter) to prevent overfitting. These perturbations effectively enhance the robustness of identity preservation.

\noindent\textbf{Training loss.}\quad
Following DreamVideo-2~\cite{wei2024dreamvideo2}, we use a reweighted diffusion loss that differentiates the contributions of regions inside and outside the bounding boxes.
Specifically, we amplify the contributions within bounding boxes to enhance subject learning while preserving the original diffusion loss for regions outside these boxes.
The training loss of stage 1 is defined as:
\begin{align}
\mathcal{L}_\text{sft} = \mathbb{E}_{\bm{z}, \epsilon, \mathcal{C}, t} & \left[ (1 + \lambda_1\mathbf{M}) \cdot \big\| \epsilon - \epsilon_{\theta}(\bm{z}_t, \mathcal{C}, t) \big\|_{2}^{2} \right],
\label{eq:sft_loss}
\end{align}
where $\mathcal{C} = \{ \bm{c}_\text{txt}, \bm{z}_\text{ref}, \bm{z}_\text{box}, \bm{z}_\text{traj} \}$ is the comprehensive conditioning set. $\bm{c}_\text{txt}$ is the text prompt, $\bm{z}_\text{ref}$ is the latent codes of reference images $I_\text{ref}$, and $\bm{z}_\text{traj}$ is the constructed trajectory feature map.
$\mathbf{M}$ denotes the binary bounding box masks (1 for foreground, 0 otherwise), and $\lambda_1 > 0$ is the balancing factor.
$\epsilon \sim \mathcal{N}(\mathbf{0}, \mathbf{I})$ and $\epsilon_{\theta}$ denotes the denoising DiT network.

\subsection{Latent Identity Reinforcement Learning}
\label{sec:method_rl}
While the omni-motion and identity SFT stage establishes unified controllability, relying solely on low-level reconstruction losses is insufficient for preserving fine-grained appearance details. 
To further enhance identity fidelity by aligning with human preferences, we introduce the latent identity reinforcement learning stage, which trains a latent identity reward model for reward feedback learning, as shown in Fig.~\ref{fig:framework}.

\subsubsection{Latent Identity Reward Model}
To provide fine-grained, identity-consistent feedback during reinforcement learning, we introduce the Latent Identity Reward Model (LIRM). 
Unlike conventional reward models (\eg, CLIP or VLMs) that evaluate videos in RGB space, our LIRM operates directly in latent space, mitigating computational overhead and facilitating the subsequent reward feedback learning.

\noindent\textbf{Architecture.}\quad
Fig.~\ref{fig:framework} shows that LIRM comprises three key modules: a video diffusion model (VDM) backbone, an identity cross-attention layer, and a reward prediction head. 
Specifically, given a video pair $V \in \{V_{pos}, V_{neg}\}$ and a reference image $I_\text{ref}$, we first project them into a shared latent space via a 3D VAE encoder, yielding latent representations $\bm{z}_V$ and $\bm{z}_\text{ref}$.
We perturb $\bm{z}_V$ with gaussian noise at timestep $t$ to yield $\bm{z}_{V,t}$, while maintaining $\bm{z}_\text{ref}$ in its clean state.
We then leverage the pretrained VDM backbone $\Phi$ to extract spatiotemporal features $\bm{f}_V = \Phi(\bm{z}_{V,t}, t, \bm{c}_\text{txt})$ from the noisy video and identity features $\bm{f}_\text{ref} = \Phi(\bm{z}_\text{ref}, t_0, \bm{c}_\text{txt})$ from the reference image.
Subsequently, the identity features serve as the query $\mathbf{Q}$ in a cross-attention layer to attend to the video's spatiotemporal features acting as key $\mathbf{K}$ and value $\mathbf{V}$, measuring the alignment between the subject's identity and the generated video content:
\begin{equation}
    \mathbf{h}_{\text{attn}} = \operatorname{Attention}(\mathbf{Q}, \mathbf{K}, \mathbf{V}) = \operatorname{Softmax}\left(\frac{\mathbf{Q}\mathbf{K}^\top}{\sqrt{d}}\right)\mathbf{V},
\end{equation}
where $\mathbf{Q} = \bm{f}_\text{ref}\mathbf{W}_\mathbf{Q}$, and $\mathbf{K}, \mathbf{V} = \bm{f}_V\mathbf{W}_\mathbf{K}, \bm{f}_V\mathbf{W}_\mathbf{V}$.
Finally, a residual connection fuses the aligned features with the query, and the resulting representation is passed through a lightweight MLP head $\mathcal{H}$ to predict the scalar reward $r_t$:
\begin{equation}
    r_t = \mathcal{H}(\mathbf{h}_{\text{attn}} + \mathbf{Q}).
\end{equation}
Unlike prior works~\cite{meng2025identity, shen2025identity} that rely on static image-based encoders as reward models, our LIRM leverages the inherent spatiotemporal priors of the VDM. 
This facilitates motion-aware identity assessment by evaluating identity consistency integrated with motion dynamics, penalizing ``copy-paste'' artifacts while ensuring robust preservation under large motion.

\noindent\textbf{Latent identity preference optimization.}\quad 
We curate a high-quality preference dataset $\mathcal{D}_\text{LIRM} = \{(V, I_\text{ref}, \bm{c}_\text{txt}, y)_i\}_{i=1}^N$ from our in-house data, comprising $\sim$27,500 training videos and 500 testing videos. Each sample consists of video win-lose pairs coupled with corresponding single- or multi-subject reference images.
Each video is assigned a human-annotated label $y \in \{0, 1\}$ to indicate whether the video $V$ aligns with the identity defined by 
$I_\text{ref}$.
Leveraging this dataset, we optimize the LIRM parameters via a binary cross-entropy loss:
\begin{equation}
    \mathcal{L}_\text{LIRM} = - \mathbb{E}_{\mathcal{D}_\text{LIRM}} \big[ y \log \sigma(r_t) + (1 - y) \log (1 - \sigma(r_t)) \big],
\end{equation}
where $\sigma(\cdot)$ is the sigmoid activation.
To mitigate computational overhead, we utilize the first eight blocks of the VDM as the backbone, following~\cite{mi2025video}. 
During training, the VDM backbone, identity cross-attention layer, and reward prediction head are jointly updated.

\subsubsection{Latent Identity Reward Feedback Learning}
Benefiting from the architectural efficiency of our LIRM, we perform Reward Feedback Learning (ReFL) within the latent space to further enhance identity preservation by aligning with human preferences.
Standard ReFL faces severe computational bottlenecks in video generation, as it necessitates expensive VAE decoding for pixel-level evaluation. Furthermore, it typically restricts reward feedback to the final denoised result, thereby neglecting structural information established in the early diffusion stages.
In contrast, by bypassing the VAE decoder, our Latent Identity Reward Feedback Learning (LIReFL) significantly mitigates memory overhead. This design enables direct gradient backpropagation to the video generator and dense reward feedback at arbitrary diffusion timesteps, thereby fully leveraging the potential of ReFL.

Specifically, we initialize the latents from Gaussian noise and sample a target intermediate timestep $t_m \sim \mathcal{U}(0, T-1)$.
We first perform standard gradient-free denoising from step $T$ down to $t_{m+1}$ to conserve memory. 
Upon reaching step $t_{m+1}$, we execute a single gradient-enabled denoising step to derive the predicted latent $\bm{z}_{t_m}$, which is formulated as:
\begin{equation}
\bm{z}_{t_m} = \mu_{\theta}(\bm{z}_{t_{m+1}}, t_{m+1}, \bm{c}_\text{txt}, \bm{z}_\text{ref}),
\end{equation}
where $\mu_{\theta}$ denotes the single-step solver function (\eg, UniPC~\cite{zhao2023unipc} step) parameterized by the video generator $\epsilon_{\theta}$.
The resulting $\bm{z}_{t_m}$ is immediately evaluated by the frozen LIRM to predict the identity reward $r_{t_m} = \text{LIRM}(\bm{z}_{t_m}, t_{m}, \bm{c}_\text{txt}, \bm{z}_\text{ref})$.
The reinforcement loss is then formulated to maximize this expected identity fidelity:
\begin{equation}
\mathcal{L}_\text{LIReFL} = -\mathbb{E}_{t_{m}, \bm{c}_\text{txt}, \bm{z}_\text{ref}} [r_{t_m}].
\end{equation}

To prevent ``reward hacking'', where the model over-optimizes for identity scores at the expense of visual quality or diversity, we incorporate the supervised SFT objective from the first stage as a regularizer. The final training objective is a weighted combination:
\begin{equation}
\mathcal{L} = \mathcal{L}_\text{sft} + \lambda_2 \mathcal{L}_\text{LIReFL},
\end{equation}
where $\lambda_2$ controls the strength of the reward feedback.
This balanced strategy ensures the model aligns with human identity preferences while preserving precise motion control and generative diversity established during the SFT stage.

\begin{figure}[t]
  \centering
  \includegraphics[width=1.0\linewidth]{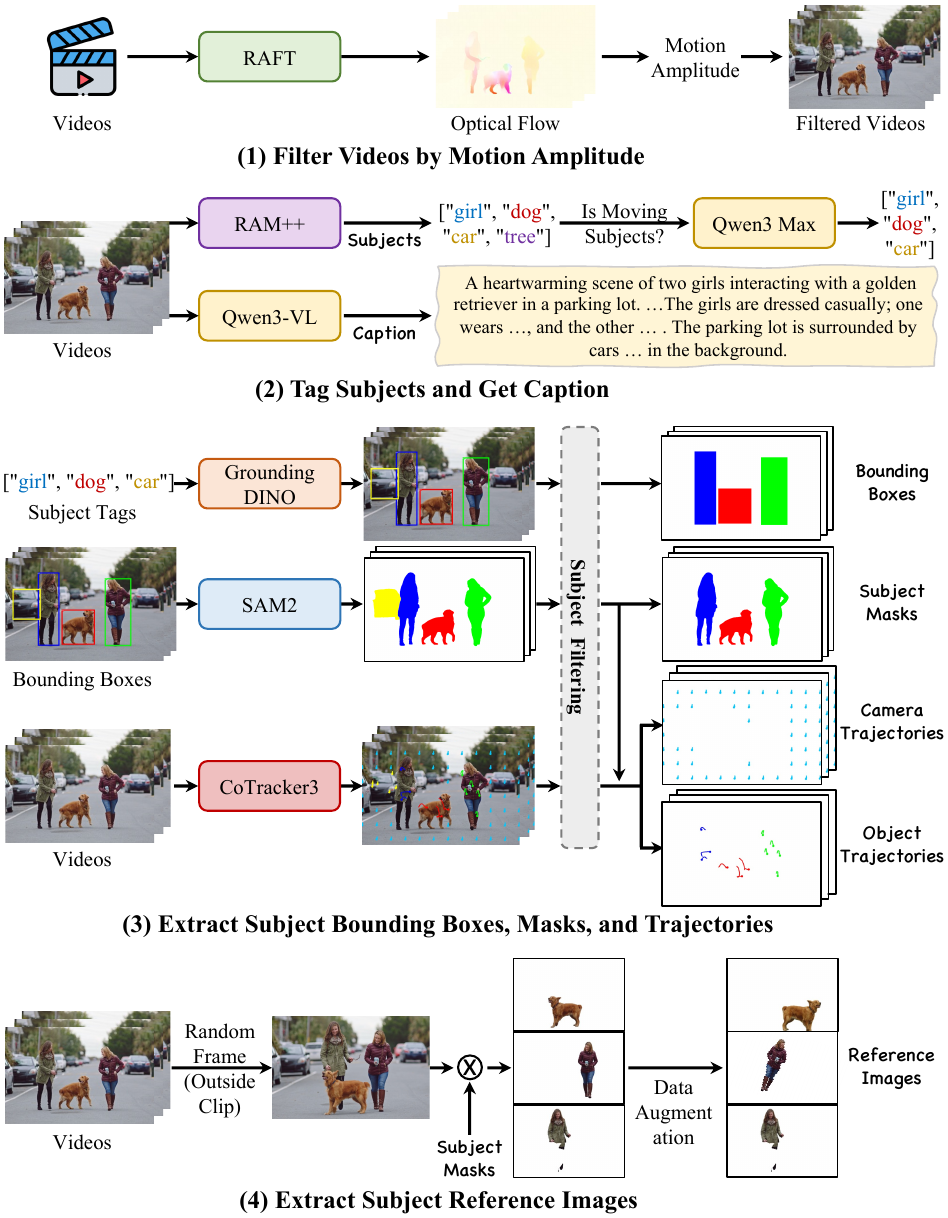}
  \caption{\textbf{Pipeline of dataset construction}.
  }
  \label{fig:dataset_construction}
\end{figure}

\subsection{Dataset Construction Pipeline}
\label{sec:data_construction}
To facilitate the SFT stage of \frameworkplain, which demands precise alignment across subject identity, global and local motion, and camera movement, we construct a large-scale, densely annotated, high-quality video dataset. Fig.~\ref{fig:dataset_construction} illustrates our automated pipeline with four sequential stages:

\noindent\textbf{1) Motion-based filtering.}\quad
Robust motion control learning necessitates training samples with significant temporal dynamics. We estimate dense optical flow using RAFT~\cite{teed2020raft} and compute the average motion magnitude across frames. Videos with small motion magnitude are discarded, ensuring the dataset focuses on meaningful motion patterns.

\noindent\textbf{2) Subject discovery and captioning.}\quad
To identify moving subjects and generate dense descriptions, we first utilize RAM++~\cite{zhang2024recognize} to extract semantic tags from the video. 
These tags are subsequently refined by Qwen3 Max~\cite{yang2025qwen3} to retain only significant moving subjects. 
Finally, we employ Qwen3-VL~\cite{Qwen3-VL} to generate detailed captions for each video.

\noindent\textbf{3) Spatiotemporal annotation extraction.}\quad
This stage extracts core structural conditions, including global bounding boxes, subject masks, and motion trajectories. 
Guided by the filtered tags, we first employ Grounding DINO~\cite{liu2024grounding} to detect subject bounding boxes, which serve as inputs for SAM 2~\cite{ravi2024sam2} to yield precise binary segmentation masks of subjects. 
Then, we utilize CoTracker3~\cite{karaev2025cotracker3} for dense point tracking and classify the resulting trajectories based on the subject masks: points falling within subject regions are labeled as object trajectories, while those in the background are designated as camera trajectories.

\noindent\textbf{4) Reference image construction.}\quad
To facilitate zero-shot customization and mitigate trivial copy-paste solutions, we sample reference images from frames temporally disjoint from the training clip. 
The subjects are then isolated via their corresponding segmentation masks and applied to extensive data augmentation to yield final reference images.

Table~\ref{tab:dataset_compare} shows that distinct from previous video customization and controllable generation datasets, our dataset uniquely supports multi-subject customization with comprehensive motion annotations, including segmentation masks, bounding boxes, and trajectories. 
This richly annotated corpus establishes a solid foundation for both video customization and motion control tasks.

\begin{table}[t]
    \caption{\textbf{Comparison with datasets for video customization and controllable generation.} Our dataset uniquely supports multi-subject customization with comprehensive motion control annotations.}
    \centering
    \footnotesize
    \setlength\tabcolsep{1.5pt}
    \resizebox{\columnwidth}{!}{
    \begin{tabular}{lcccccc}
        \toprule
        & 
        \tabincell{c}{\textbf{No. of}\\\textbf{Videos}} & 
        \tabincell{c}{\textbf{Reference}\\\textbf{Images}} & 
        \tabincell{c}{\textbf{Multi-}\\\textbf{Subject}} &   
        \tabincell{c}{\textbf{All-Frame}\\\textbf{Mask}} &   
        \tabincell{c}{\textbf{All-Frame}\\\textbf{Box}} &    
        \tabincell{c}{\textbf{All-Frame}\\\textbf{Trajectory}} 
        \\ 
        \midrule
        WebVid-10M~\cite{webvid10m} & $\sim$10M & \xmark & \xmark & \xmark & \xmark & \xmark \\
        UCF-101~\cite{ucf101} & 13,320 & \xmark & \xmark & \xmark & \xmark & \xmark \\
        DAVIS~\cite{davis} & 50 & \xmark & \xmark & \cmark & \cmark & \xmark \\
        GOT-10k~\cite{GOT10k} & 9,695 & \xmark & \xmark & \xmark & \cmark & \xmark \\
        VideoBooth~\cite{jiang2024videobooth} & 48,724 & \cmark & \xmark & \xmark & \xmark & \xmark \\
        DreamVideo-2~\cite{wei2024dreamvideo2} & 230,160 & \cmark & \xmark & \cmark & \cmark & \xmark \\
        Video Alchemist~\cite{chen2025multi} & $\sim$37.8M & \cmark & \cmark & \xmark & \xmark & \xmark \\
        Phantom~\cite{liu2025phantom} & $\sim$1M & \cmark & \cmark & \xmark & \xmark & \xmark \\
        Wan-Move~\cite{chu2025wan} & $\sim$1.98M & \xmark & \xmark & \xmark & \xmark & \cmark \\
        \midrule
        \textbf{Our Dataset} & \textbf{$\sim$2.12M} & \cmark & \cmark & \cmark & \cmark & \cmark \\ 
        \bottomrule
    \end{tabular}
    }
    \label{tab:dataset_compare}
\end{table}

\begin{figure}[t]
  \centering
  \includegraphics[width=0.9\linewidth]{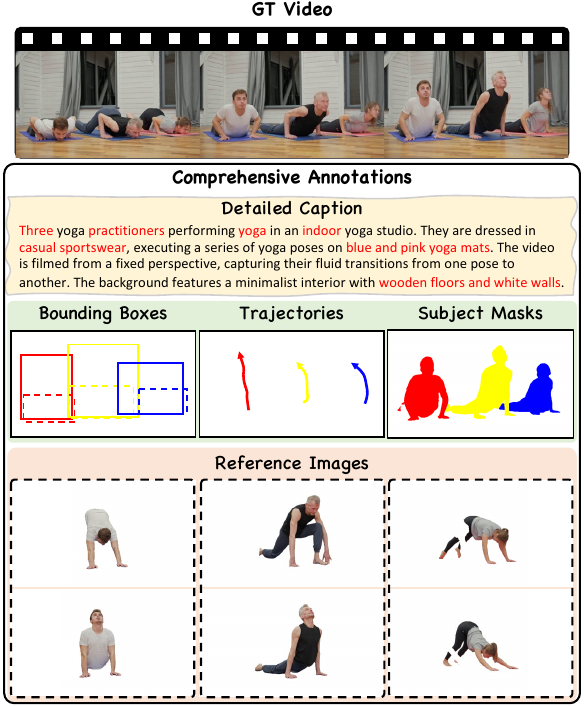}
  \caption{\textbf{Visualization of a test sample from DreamOmni Bench.} Our benchmark supports fine-grained evaluation through comprehensive annotations, including multiple reference images for each subject, detailed captions, and precise spatial-temporal ground truths such as bounding boxes, motion trajectories, and subject masks.
}
  \label{fig:benchmark_sample}
\end{figure}

\subsection{DreamOmni Bench}
\label{sec:benchmark}
Existing benchmarks typically isolate video customization from controllable generation, rendering them inadequate for evaluating the holistic capabilities of \frameworkplain. 
On the one hand, current personalization benchmarks~\cite{wei2024dreamvideo, wei2024dreamvideo2, jiang2024videobooth} are predominantly confined to single-subject scenarios, lacking the capacity to assess multi-subject consistency and quantify motion controllability. 
On the other hand, recent motion control benchmarks, such as Wan-Move~\cite{chu2025wan}, focus exclusively on point trajectory precision. These protocols are neither comprehensive nor capable of measuring identity preservation.
Consequently, there is a critical absence of a unified benchmark that \textit{simultaneously evaluates} multi-subject customization and comprehensive motion control, including bounding boxes and dense point trajectories.

To bridge this gap, we construct the \textbf{DreamOmni Bench}, composed of high-quality real-world videos sourced independently of our training dataset to ensure zero-shot evaluation.
We perform manual filtering to retain high-resolution videos that exhibit meaningful subject motion and camera movement, explicitly excluding static videos and frames with text overlays or watermarks.
After filtering, we leverage the automated pipeline detailed in Sec.~\ref{sec:data_construction} to generate dense captions and comprehensive annotations (including subject masks, bounding boxes, and trajectories) for each video.
The resulting benchmark comprises a total of 1,027 videos, explicitly categorized into 436 single-subject and 591 multi-subject samples, featuring diverse categories spanning humans, general objects, animals, and faces.

Our DreamOmni Bench facilitates a unified evaluation of both identity preservation (covering both generic objects and human faces) and motion control precision (measuring bounding box and trajectory accuracy). 
Specific quantitative metrics are detailed in Sec.~\ref{sec:exp_setup}, while visual samples from the benchmark are provided in Fig.~\ref{fig:benchmark_sample}.

\section{Experiment}
\label{sec:exp}
\subsection{Experimental Setup}
\label{sec:exp_setup}

\noindent\textbf{Implementation details.}\quad
We utilize Wan2.1-1.3B T2V as our foundational model. Across all training stages, we employ the AdamW optimizer~\cite{adamw} to process video clips at a resolution of $480 \times 832$ with $49$ frames.
\textbf{1)} In the first omni-motion and identity SFT stage, the model is fine-tuned for 40,000 iterations on 64 NVIDIA A100 GPUs with total batch size $64$, using a learning rate of $5 \times 10^{-5}$ and weight decay of $1 \times 10^{-3}$.
To enhance robustness, we randomly drop bounding box and trajectory conditions with a probability of $p=0.5$, and apply data augmentation to reference images with the same probability ($p=0.5$). The reweighted diffusion loss weight $\lambda_1$ is set to $2$.
\textbf{2)} The second latent identity reinforcement learning stage comprises two sub-steps: Latent Identity Reward Model (LIRM) training and Latent Identity Reward Feedback Learning (LIReFL). Both are conducted on 16 A100 GPUs with a batch size of 16 and a weight decay of $1 \times 10^{-2}$.
For LIRM training, we initialize the backbone using the first 8 layers of Wan2.1-1.3B and train for $\sim$4,000 steps. We use differential learning rates: $1 \times 10^{-5}$ for the prediction head and attention layer, and $1 \times 10^{-6}$ for the VDM backbone. During training, we freeze the text and patch embedding layers of the pretrained VDM.
For LIReFL, we fine-tune the DiT from the SFT stage for 3,400 steps while keeping the reward model frozen. We incorporate an SFT loss as a regularizer with a weight of $\lambda_2=0.1$, and set the learning rate to $5 \times 10^{-6}$. The condition dropping and reference augmentation strategies follow the same protocol as the SFT stage.
During inference, we employ the UniPC scheduler~\cite{zhao2023unipc} with 50 steps and a classifier-free guidance scale~\cite{ho2022classifier_free_guide} of 5.0.

\noindent\textbf{Baselines.}\quad
Due to the absence of open-source methods capable of simultaneously supporting multi-subject customization and comprehensive motion control, we benchmark \frameworkplain against prior methods from three distinct categories on both the DreamOmni Bench and MSRVTT-Personalization Bench~\cite{chen2025multi}.
On the DreamOmni Bench, we compare with:
(1) DreamVideo-2~\cite{wei2024dreamvideo2}, representing single-subject customization with motion control;
(2) VACE~\cite{jiang2025vace} and Phantom~\cite{liu2025phantom}, focusing on single and multi-subject customization; and
(3) Wan-Move~\cite{chu2025wan}, specializing in trajectory control.
Additionally, on the MSRVTT-Personalization Bench, we compare with the recent state-of-the-art methods Video Alchemist~\cite{chen2025multi} and Tora2~\cite{zhang2025tora2}. Since their codes are not publicly available, we cite the quantitative results directly from their original papers.

\begin{figure*}[t]
  \centering
  \includegraphics[width=1.0\linewidth]{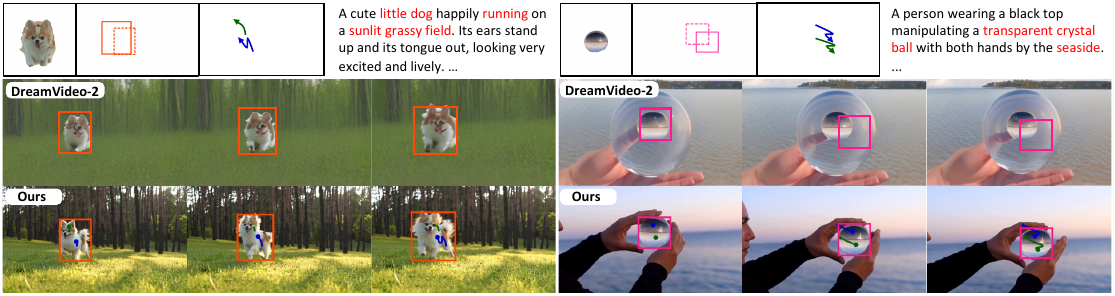}
  \caption{\textbf{Qualitative comparison of joint subject customization and motion control.} Previous methods struggle to balance identity preservation with accurate motion control. In contrast, our method delivers high-fidelity subject customization that strictly follows complex spatial trajectories.}
  \label{fig:compare_id_motion}
\end{figure*}
\begin{table*}[t]
    \centering
    \caption{\textbf{Quantitative comparison on DreamOmni Bench.}}
    \label{tab:dreamomni_comparison}
    \begin{tabular}{lcccccc} 
        \toprule
        \textbf{Method} & \textbf{R-CLIP} $\uparrow$ & \textbf{R-DINO} $\uparrow$ & \textbf{Face-S} $\uparrow$ & \textbf{mIoU} $\uparrow$ & \textbf{EPE} $\downarrow$ & \textbf{CLIP-T} $\uparrow$ \\ 
        \midrule
        DreamVideo-2~\cite{wei2024dreamvideo2} & 0.731 & 0.429 & 0.157 & 0.212 & 24.05 & 0.297 \\
        \textbf{\framework} (Ours) & \textbf{0.739} & \textbf{0.499} & \textbf{0.301} & \textbf{0.558} & \textbf{9.31} & \textbf{0.308} \\
        \bottomrule
    \end{tabular}
\end{table*}

\begin{table*}[t]
    \centering
    \caption{\textbf{Quantitative comparison on MSRVTT-Personalization Bench.} We follow the experimental settings and evaluation protocols of Tora2~\cite{zhang2025tora2} and Video Alchemist~\cite{chen2025multi}, reporting results from their original papers.}
    \setlength\tabcolsep{8pt}
    \begin{tabular}{lcccccc} 
        \toprule
        \multirow{2}{*}{\textbf{Method}} & \multicolumn{3}{c}{\textbf{Subject Mode}} & \multicolumn{3}{c}{\textbf{Face Mode}} \\
        \cmidrule(lr){2-4} \cmidrule(lr){5-7}
        & \textbf{CLIP-T} $\uparrow$ & \textbf{R-DINO} $\uparrow$ & \textbf{EPE} $\downarrow$ & \textbf{CLIP-T} $\uparrow$ & \textbf{Face-S} $\uparrow$ & \textbf{EPE} $\downarrow$ \\ 
        \midrule
        Tora + Flux.1~\cite{zhang2025tora} & 0.254 & 0.587 & 19.72 & 0.265 & 0.363 & 17.41 \\
        Video Alchemist~\cite{chen2025multi}  & \underline{0.268} & \underline{0.626} & -     & 0.272 & 0.411 & -     \\
        Tora2~\cite{zhang2025tora2} & \textbf{0.273} & 0.615 & 17.43 & \textbf{0.274} & \textbf{0.419} & 13.52 \\
        \textbf{\framework} (Ours) & \textbf{0.273} & \textbf{0.628} & \textbf{11.21} & \underline{0.273} &  \underline{0.417} & \textbf{8.50} \\
        \bottomrule
    \end{tabular}
    \label{tab:msrvtt_comparison}
\end{table*}

\noindent\textbf{Evaluation metrics.}\quad
We quantitatively evaluate our method using 6 metrics across three comprehensive dimensions:
\textbf{1) Overall Consistency.} 
To assess the semantic alignment between generated videos and text prompts, we employ the CLIP-Text similarity (CLIP-T) using the CLIP ViT-B/32~\cite{clip}.
\textbf{2) Subject and Face Fidelity.} 
Whole-image similarity metrics are inadequate for multi-subject customization, as background noise and other subjects interfere with accurate identity extraction. To address this, we adopt region-based metrics to evaluate both subject and face identity preservation, specifically Region CLIP-Image similarity (R-CLIP), Region DINO-Image similarity (R-DINO), and Face Similarity (Face-S).
For R-CLIP and R-DINO, we utilize GroundingDINO~\cite{liu2024grounding} to detect and crop subject regions in generated frames based on their textual tags. We then compute the cosine similarity between the cropped regions and reference images using CLIP ViT-B/32 and DINO-ViT-S/16~\cite{dino}, respectively.
For Face-S, we employ the InsightFace library with ArcFace~\cite{deng2019arcface} for identity verification. To handle multi-person scenarios, we detect all faces in the generated frames and extract their embeddings; subsequently, we compute the cosine similarity between each detected face and the reference face, matching the generated face with the highest similarity to the ground truth for evaluation.
\textbf{3) Motion Control Precision.} 
We employ Mean Intersection over Union (mIoU) and End Point Error (EPE) to measure the accuracy of spatial layout and trajectory control.
For mIoU, we detect subjects in the generated videos using GroundingDINO and calculate the overlap between the detected bounding boxes and the ground-truth control boxes. For fine-grained trajectory evaluation (EPE), we initialize query points using the ground-truth coordinates from the first frame. These points are tracked in the generated video using CoTracker3~\cite{karaev2025cotracker3}, and EPE is computed as the average Euclidean distance between tracked and ground-truth trajectories.

\begin{figure*}[t]
  \centering
  \includegraphics[width=1.0\linewidth]{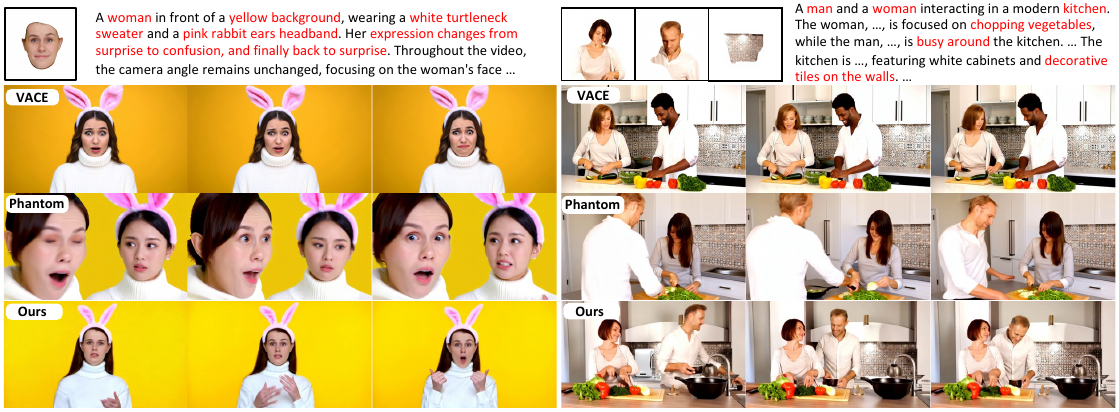}
  \caption{\textbf{Qualitative comparison of subject customization}. 
  \frameworkplain generates videos with accurate subject appearance and enhanced motion dynamics, aligning with provided prompts.
  }
  \label{fig:compare_id}
\end{figure*}

\subsection{Main Results}
\label{sec:main_results}

\noindent\textbf{Subject customization with omni-motion control.}\quad
To evaluate the capability of our framework in joint subject customization and motion control, we benchmark \frameworkplain against DreamVideo-2~\cite{wei2024dreamvideo2}, a representative baseline in this domain.
It is worth noting that while recent works like Tora2~\cite{zhang2025tora2} explore this task, their code remains closed-source, precluding direct comparison.
Furthermore, DreamVideo-2 is inherently limited to single-subject customization with coarse bounding box control.
In contrast, our \frameworkplain supports a more versatile setting, enabling multi-subject customization with omni-motion control (combining boxes and trajectories).
For a fair comparison, we restrict our evaluation to single-subject scenarios compatible with the baselines.

Tables~\ref{tab:dreamomni_comparison} and~\ref{tab:msrvtt_comparison} present the quantitative results on DreamOmni Bench and MSRVTT-Personalization Bench~\cite{chen2025multi}, respectively.
As observed, \frameworkplain significantly outperforms DreamVideo-2 across all metrics on DreamOmni Bench, demonstrating our superior subject customization and precise motion control capabilities.
Results on MSRVTT-Personalization further validate our robustness: in Subject Mode, we achieve the highest R-DINO and best EPE scores.
For Face Mode, while our Face-S score is comparable to Tora2 due to the limited video quality and resolution of the MSRVTT dataset, we still achieve significantly better EPE.
These consistent improvements across diverse benchmarks underscore the generalization of \frameworkplain in delivering high-fidelity subject customization and precise motion control.

Fig.~\ref{fig:compare_id_motion} demonstrates the robustness of \frameworkplain across diverse scenarios.
Fig.~\ref{fig:compare_id_motion} (left) shows that while our method maintains high identity fidelity, DreamVideo-2 suffers from identity degradation under large motion.
Conversely, for complex trajectories shown on the right, DreamVideo-2 often exhibits trajectory drift or static motion, failing to achieve precise motion control, while our method precisely aligns with the bounding boxes and trajectories.
These results reveal that DreamVideo-2 struggles to balance identity preservation and motion control.
In contrast, our framework effectively resolves this conflict, simultaneously achieving precise motion control and high-fidelity subject customization.

\noindent\textbf{Subject customization.}\quad
We evaluate the pure subject customization capability of \frameworkplain on the DreamOmni Bench, comparing it against state-of-the-art methods including VACE~\cite{jiang2025vace} and Phantom~\cite{liu2025phantom} across both single-subject and multi-subject scenarios.

Table~\ref{tab:customization_compare} shows that our method achieves state-of-the-art performance on single-subject customization.
Specifically, \frameworkplain yields the highest R-DINO and R-CLIP scores, indicating superior identity preservation.
In the more challenging multi-subject setting, our method consistently surpasses baselines in R-DINO, Face-S, and CLIP-T scores.
This demonstrates that our design effectively prevents identity mixing and leakage while achieving superior text alignment.

Fig.~\ref{fig:compare_id} (left) illustrates the results in a single-subject scenario where the subject reveals her face from behind a leaf.
VACE exhibits suboptimal facial preservation and unnatural motion, while Phantom generates unexpected multiple faces.
In contrast, our method generates a natural transition with the face faithfully preserved.
In the multi-subject scenario involving a complex interaction between a man and a woman, both baselines fail to preserve identity fidelity, exhibiting discrepancies in skin tone, hairstyle, facial details, and clothing color compared to the reference.
Conversely, \frameworkplain successfully disentangles the two identities, rendering distinct, correct appearances with coherent interactions.
This validates the effectiveness of our approach in maintaining high-fidelity identity, even in challenging scenarios.

\begin{table}[t]
    \centering
    \caption{\textbf{Quantitative comparison of subject customization on DreamOmni Bench.} We report results separately for single-subject and multi-subject scenarios. All models have 1.3B parameters.}
    \setlength\tabcolsep{4pt}
    \begin{tabular}{lcccc} 
        \toprule
        \textbf{Method} & \textbf{R-CLIP} $\uparrow$ & \textbf{R-DINO} $\uparrow$ & \textbf{Face-S} $\uparrow$ & \textbf{CLIP-T} $\uparrow$ \\ 
        \midrule
        \multicolumn{5}{c}{\textit{\textbf{Single-Subject Mode}}} \\
        \midrule
        VACE~\cite{jiang2025vace} & 0.732 & 0.480 & 0.174 & 0.293 \\
        Phantom~\cite{liu2025phantom} & 0.738 & 0.485 & 0.299 & 0.296 \\
        \textbf{\framework}    & \textbf{0.739} & \textbf{0.499} & \textbf{0.301} & \textbf{0.308} \\
        \midrule
        \multicolumn{5}{c}{\textit{\textbf{Multi-Subject Mode}}} \\
        \midrule
        VACE~\cite{jiang2025vace} & 0.719 & 0.497 & 0.275 & 0.293 \\
        Phantom~\cite{liu2025phantom} & \textbf{0.722} & 0.517 & 0.305 & 0.293 \\
        \textbf{\framework} & 0.720 & \textbf{0.524} & \textbf{0.329} & \textbf{0.306}\\
        \bottomrule
    \end{tabular}
    \label{tab:customization_compare}
\end{table}
\begin{table}[t]
    \centering
    \caption{\textbf{Quantitative comparison of motion control capability on DreamOmni Bench.} We report results separately for single-subject and multi-subject scenarios.}
    \footnotesize
    \begin{tabular}{lccc} 
        \toprule
        \textbf{Method} & \textbf{mIoU} $\uparrow$ & \textbf{EPE} $\downarrow$ & \textbf{CLIP-T} $\uparrow$ \\ 
        \midrule
        \multicolumn{4}{c}{\textit{\textbf{Single-Subject Mode}}} \\
        \midrule
        Tora~\cite{zhang2025tora} (T2V, 1.1B) & 0.163 & 31.74 & 0.307 \\
        Wan-Move~\cite{chu2025wan} (I2V, 14B) & 0.507 & 14.43 & 0.305 \\
        \textbf{\framework} (T2V, 1.3B) & \textbf{0.558} & \textbf{9.31} & \textbf{0.308} \\
        \midrule
        \multicolumn{4}{c}{\textit{\textbf{Multi-Subject Mode}}} \\
        \midrule
        Tora~\cite{zhang2025tora} (T2V, 1.1B) & 0.162 & 32.84 & \textbf{0.306} \\
        Wan-Move~\cite{chu2025wan} (I2V, 14B) & 0.541 & 9.02 & 0.303 \\
        \textbf{\framework} (T2V, 1.3B) & \textbf{0.570} & \textbf{6.08} & \textbf{0.306} \\
        \bottomrule
    \end{tabular}
    \label{tab:motion_compare}
\end{table}

\begin{figure*}[t]
  \centering
  \includegraphics[width=1.0\linewidth]{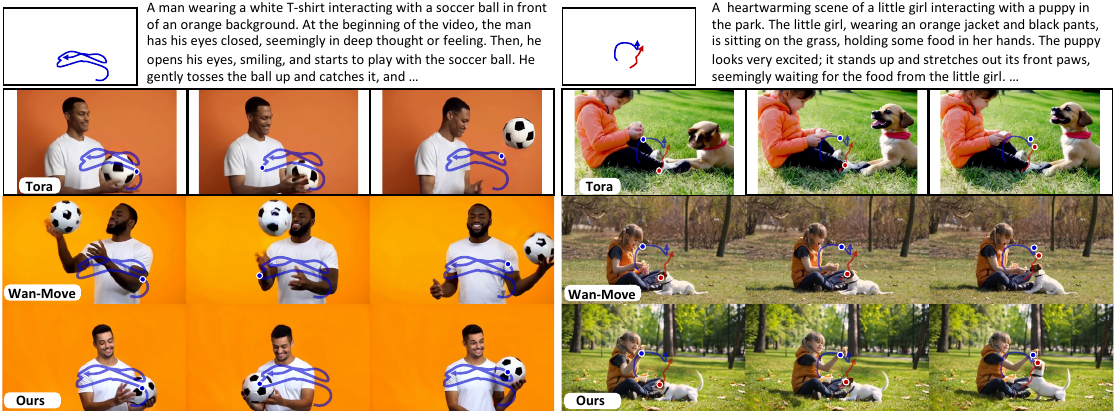}
  \caption{\textbf{Qualitative comparison of motion control}.
  Our \frameworkplain achieves precise motion trajectory control.}
  \label{fig:compare_motion}
\end{figure*}

\noindent\textbf{Motion control.}\quad
To validate the effectiveness of our motion control capabilities, we compare \frameworkplain against state-of-the-art models, including the Tora~\cite{zhang2025tora} (T2V, 1.1B) and the large-scale Wan-Move~\cite{chu2025wan} (I2V, 14B).

Table~\ref{tab:motion_compare} shows that \frameworkplain consistently outperforms baselines across both single-subject and multi-subject scenarios.
Compared to Tora, our method achieves a substantial improvement in motion precision, with mIoU increasing by 0.395 and EPE reducing by nearly 70\%.
Notably, despite being a 1.3B parameter model, \frameworkplain surpasses the 14B parameter Wan-Move across all metrics in both settings, demonstrating the significant parameter efficiency and superior controllability of our method.

Fig.~\ref{fig:compare_motion} further highlights these differences.
Tora struggles with trajectory adherence, failing to maintain effective control in multi-subject settings.
While Wan-Move generates high-quality visual content, it tends to deviate from complex trajectories, as evidenced by the inaccurate path of the soccer ball and the loose alignment between the puppy and the girl's hand movements.
In contrast, \frameworkplain precisely follows complex motion trajectories.
In the single-subject case, the man's interaction with the soccer ball strictly adheres to the intricate looping trajectory.
In the multi-subject scenario, our model accurately coordinates the distinct movements of the girl and the puppy, verifying our robust control capabilities.

\noindent\textbf{User study.}\quad
To further evaluate the perceptual quality of generated videos, we conduct a comprehensive user study focusing on three distinct capabilities: joint subject customization with motion control, pure subject customization, and pure motion control.
We invite 18 evaluators to rate 270 groups of videos generated by different methods.
Each evaluation group consists of reference subject images, a target textual prompt, corresponding motion conditions (\ie, bounding boxes or trajectories), and the videos generated by competing methods.
Evaluators are asked to select the best video based on four criteria: Subject Fidelity, Motion Consistency, Text Alignment, and Overall Quality.
Table~\ref{tab:user_study} shows that our method achieves the highest user preference across diverse settings.

\begin{table}[t]
    \centering
    \small
    \caption{\textbf{Human Evaluation.} We report the percentage of user votes for each method across different settings.}
    \label{tab:user_study}
    \resizebox{\linewidth}{!}{
        \begin{tabular}{l|l|cccc}
            \toprule
            \textbf{Setting} & \textbf{Method} & \textbf{\tabincell{c}{Subject \\ Fidelity}} & \textbf{\tabincell{c}{Motion \\ Consistency}} & \textbf{\tabincell{c}{Text \\ Alignment}} & \textbf{\tabincell{c}{Overall \\ Quality}} \\
            \midrule
            \multirow{2}{*}{\tabincell{l}{Joint ID \& \\ Motion}} 
            & DreamVideo-2~\cite{wei2024dreamvideo2} & 22.4\% & 18.3\% & 21.5\% & 10.8\% \\
            & \textbf{Ours} & \textbf{77.6\%} & \textbf{81.7\%} & \textbf{78.5\%} & \textbf{89.2\%} \\
            \midrule
            \multirow{3}{*}{\tabincell{l}{Pure Subject \\ Customization}} 
            & VACE~\cite{jiang2025vace} & 16.3\% & - & 15.6\% & 19.5\% \\
            & Phantom~\cite{liu2025phantom} & 19.5\% & - & 16.8\% & 20.2\% \\
            & \textbf{Ours} & \textbf{64.2\%} & - & \textbf{67.6\%} & \textbf{60.3\%} \\
            \midrule
            \multirow{3}{*}{\tabincell{l}{Pure Motion \\ Control}} 
            & Tora~\cite{zhang2025tora} & - & 9.5\% & 16.5\% & 13.4\% \\
            & Wan-Move~\cite{chu2025wan} & - & 20.2\% & 20.4\% & 26.4\% \\
            & \textbf{Ours} & - & \textbf{70.3\%} & \textbf{63.1\%} & \textbf{60.2\%} \\
            \bottomrule
        \end{tabular}
    }
\end{table}
\begin{figure*}[t]
  \centering
  \includegraphics[width=1.0\linewidth]{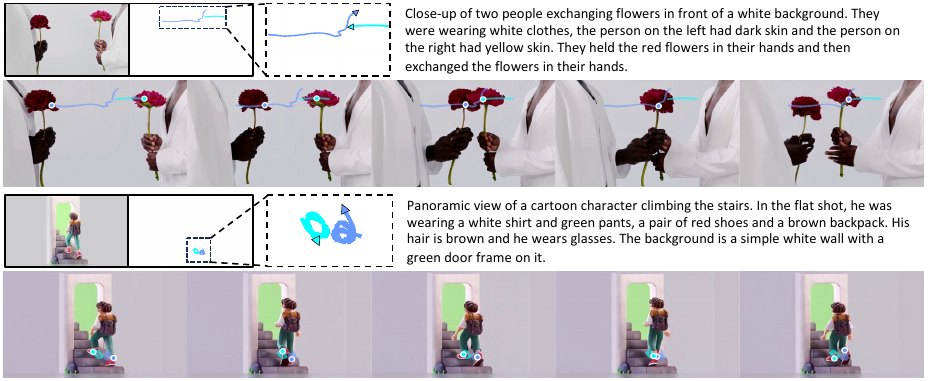}
  \caption{\textbf{Emergent generative capabilities of \frameworkplain.} Despite being built on a text-to-video (T2V) base model, our framework naturally enables zero-shot Image-to-Video (I2V) generation and first-frame-conditioned trajectory control without task-specific fine-tuning.}
  \label{fig:I2V}
\end{figure*}

\subsection{Emergent Capabilities}
Benefiting from the robust multi-subject customization and omni-motion control capabilities of \frameworkplain, our multi-task training paradigm facilitates \textbf{the emergence of novel generative abilities: Image-to-Video (I2V) generation and first-frame-conditioned trajectory control, despite our base DiT being a text-to-video (T2V) model}, as illustrated in Fig.~\ref{fig:I2V}.
In fact, I2V generation can be considered as a specialized form of customization, where the entire first frame serves as the comprehensive reference identity. Furthermore, our versatile omni-motion mechanism seamlessly extends to this setting, enabling precise spatial trajectory guidance directly conditioned on the provided initial frame. These emergent properties demonstrate the strong generalization and unified control capacity of our framework without requiring task-specific fine-tuning.

\begin{table}[t]
    \centering
    \caption{\textbf{Quantitative ablation studies on each component in \framework.} We report results separately for single-subject and multi-subject scenarios to analyze the impact of each module.}
    \small
    \setlength\tabcolsep{2pt}
    \resizebox{\linewidth}{!}{
    \begin{tabular}{lcccccc} 
        \toprule
        \textbf{Method} & \textbf{R-CLIP}$\uparrow$ & \textbf{R-DINO}$\uparrow$ & \textbf{Face-S}$\uparrow$ & \textbf{mIoU}$\uparrow$ & \textbf{EPE}$\downarrow$ & \textbf{CLIP-T}$\uparrow$ \\ 
        \midrule
        \multicolumn{7}{c}{\textit{\textbf{Single-Subject Mode}}} \\
        \midrule
        w/o Cond-Aware 3D RoPE      & 0.625 & 0.139 & 0.039 & 0.274 & 30.22 & 0.216 \\
        w/o Group \& Role Emb.      & \underline{0.738} & 0.486 & 0.254 & 0.524 & 26.24 & \textbf{0.309} \\
        w/o Hierarchical BBox Injection         & 0.733 & \textbf{0.508} & 0.257 & 0.400 & 31.84 & 0.307 \\
        Ours (Stage1)             & 0.733 & 0.483 & 0.251 & 0.556 & 10.53 & 0.306 \\
        \midrule
        w/o LIReFL (Stage2 SFT only) & 0.735 & 0.487 & \underline{0.266} & \textbf{0.561} & \underline{10.01} & 0.307 \\
        \textbf{Ours (Full)} & \textbf{0.739} & \underline{0.499} & \textbf{0.301} & \underline{0.558} & \textbf{9.31} & \underline{0.308} \\
        \midrule
        \multicolumn{7}{c}{\textit{\textbf{Multi-Subject Mode}}} \\
        \midrule
        w/o Cond-Aware 3D RoPE      & 0.647 & 0.157 & 0.047 & 0.278 & 20.71 & 0.224 \\
        w/o Group \& Role Emb.      & 0.708 & 0.503 & 0.289 & 0.459 & 20.69 & \textbf{0.308} \\
        w/o Hierarchical BBox Injection   & 0.714 & 0.510 & 0.269 & 0.289 & 25.56 & 0.305 \\
        Ours (Stage1) & 0.713 & 0.506 & 0.287 & 0.532 & 6.80 & 0.305 \\
        \midrule
        w/o LIReFL (Stage2 SFT only) & \underline{0.715} & \underline{0.512} & \underline{0.316} & \underline{0.556} & \underline{6.29} & \underline{0.306}\\
        \textbf{Ours (Full)}  
        & \textbf{0.720} & \textbf{0.524} & \textbf{0.329} & \textbf{0.570} & \textbf{6.08} & \underline{0.306}\\
        \bottomrule
    \end{tabular}
    }
    \label{tab:ablation_sft_components}
\end{table}
\begin{figure}[t]
  \centering
  \includegraphics[width=1.0\linewidth]{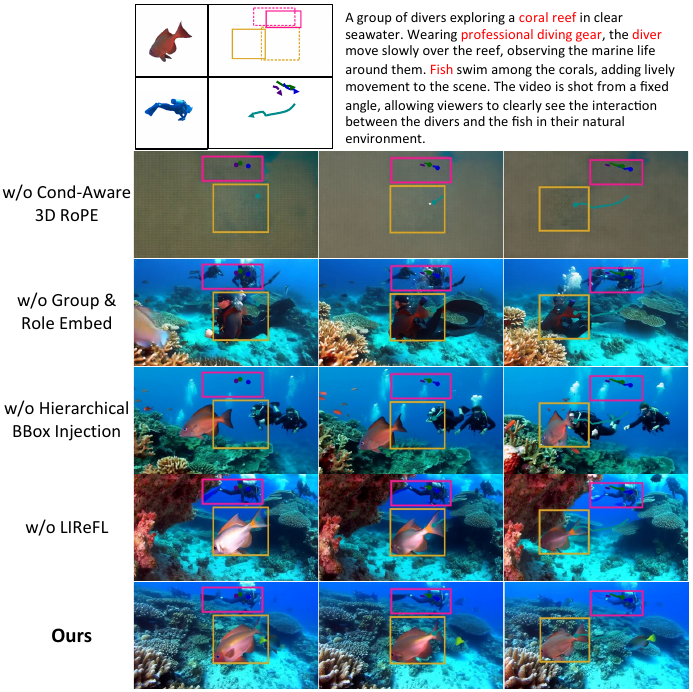}
  \caption{\textbf{Ablation study of each component in \frameworkplain}.}
  \label{fig:ablation_component}
\end{figure}

\subsection{Ablation Studies}

\noindent\textbf{Effects of each component in \framework.}\quad
To investigate the contribution of individual components, we conduct an ablation study under both single-subject and multi-subject settings on the DreamOmni Bench.

Table~\ref{tab:ablation_sft_components} reveals the critical role of each component.
\textit{(i)} Removing the condition-aware 3D RoPE leads to a catastrophic performance drop across all metrics in both scenarios, confirming its fundamental role in handling multi-condition heterogeneous inputs within the unified DiT framework.
\textit{(ii)} Ablating the group and role embeddings degrades motion control precision, resulting in inferior mIoU and EPE metrics, particularly in multi-subject modes.
\textit{(iii)} Omitting hierarchical BBox injection, where bounding box latents are added solely to the input noisy latents, leads to a significant collapse in motion performance, with mIoU dropping to 0.289 in multi-subject mode. This underscores that simple input-level fusion is insufficient and hierarchical injection is critical for effective motion control.
\textit{(iv)} Regarding the training paradigm, although Stage 1 establishes a solid baseline, subsequent fine-tuning via standard SFT (w/o LIReFL) offers limited gains. 
In contrast, our full model equipped with LIReFL achieves the best performance across most metrics, particularly in multi-subject scenarios, effectively boosting subject customization while maintaining precise motion control.

Fig.~\ref{fig:ablation_component} further corroborates these findings.
\textit{(i)} Without condition-aware 3D RoPE, the generation suffers from training collapse, resulting in severe artifacts or meaningless noise.
\textit{(ii)} The absence of group and role embeddings leads to control ambiguity, where the model struggles to disentangle multiple subjects or bind specific motions to the correct identities.
\textit{(iii)} Removing hierarchical BBox injection causes subjects to fail in adhering to bounding boxes or trajectories, demonstrating that hierarchical motion injection is essential for achieving effective motion control.
\textit{(iv)} Compared to the standard SFT (w/o LIReFL), the integration of LIReFL effectively refines identity details, delivering the harmonious balance between subject fidelity and motion control.

\begin{table}[t]
    \centering
    \caption{\textbf{Ablation studies on the latent identity reward model.} The columns denote five intervals of normalized diffusion timesteps $t \in [0, 1]$. The first row represents our default setting (BCE Loss, Ref. Image as Q, Frozen Text \& Patch Embed.), while subsequent rows illustrate the impact of altering specific components.}
    \small
    \setlength\tabcolsep{2.5pt}
    \resizebox{\linewidth}{!}{
    \begin{tabular}{lcccccc} 
        \toprule
        \textbf{Method} & \textbf{[0, 0.2]} & \textbf{(0.2, 0.4]} & \textbf{(0.4, 0.6]} & \textbf{(0.6, 0.8]} & \textbf{(0.8, 1.0]} & \textbf{Avg} \\ 
        \midrule
        \textbf{Ours (Default)} & 0.702 & 0.722 & 0.709 & 0.724 & 0.743 & \textbf{0.720} \\ 
        \midrule
        \multicolumn{7}{l}{\textit{Optimization Objective}} \\
        \quad w/ BT Loss~\cite{bradley1952rank} & 0.491 & 0.657 & 0.681 & 0.706 & 0.743 & 0.656 \\
        \midrule
        \multicolumn{7}{l}{\textit{Image Injection Strategy}} \\
        \quad w/ Ref. Image as KV & 0.451 & 0.555 & 0.415 & 0.445 & 0.408 & 0.455 \\
        \midrule
        \multicolumn{7}{l}{\textit{Parameter Tuning Scope}} \\
        \quad Tuning text \& patch embed. & 0.680 & 0.718 & 0.709 & 0.716 & 0.752 & 0.715 \\
        \bottomrule
    \end{tabular}
    }
    \label{tab:ablation_reward_model}
\end{table}

\begin{table}[t]
    \centering
    \caption{\textbf{Ablation study on the range of timestep $t_m$ in LIReFL.}}
    \label{tab:ablation_refl_reward_range}
    \small
    \setlength\tabcolsep{2pt}
    \resizebox{\linewidth}{!}{
    \begin{tabular}{lcccccc}
        \toprule
        \textbf{Range of $t_m$} & \textbf{R-CLIP}$\uparrow$ & \textbf{R-DINO}$\uparrow$ & \textbf{Face-S}$\uparrow$ & \textbf{mIoU}$\uparrow$ & \textbf{EPE}$\downarrow$ & \textbf{CLIP-T}$\uparrow$ \\ 
        \midrule
        \multicolumn{7}{c}{\textit{\textbf{Single-Subject Mode}}} \\
        \midrule
        Last 3 timesteps         & 0.737 & 0.494 & 0.293 & 0.543 & 9.98 & 0.307 \\ 
        \textbf{All timesteps (Ours)} & \textbf{0.739} & \textbf{0.499} & \textbf{0.301} & \textbf{0.558} & \textbf{9.31} & \textbf{0.308} \\ 
        \midrule
        \multicolumn{7}{c}{\textit{\textbf{Multi-Subject Mode}}} \\
        \midrule
        Last 3 timesteps         & 0.717 & 0.518 & 0.324 & \textbf{0.573} & 6.30 & \textbf{0.307} \\ 
        \textbf{All timesteps (Ours)} & \textbf{0.720} & \textbf{0.524} & \textbf{0.329} & 0.570 & \textbf{6.08} & 0.306 \\ 
        \bottomrule
    \end{tabular}
    }
\end{table}
\begin{table}[t]
    \centering
    \caption{\textbf{Ablation study on loss weight $\lambda_2$ of LIReFL.}}
    \label{tab:ablation_refl_weight}
    \small
    \setlength\tabcolsep{2pt}
    \resizebox{\linewidth}{!}{
    \begin{tabular}{lcccccc}
        \toprule
        \textbf{$\lambda_2$} & \textbf{R-CLIP}$\uparrow$ & \textbf{R-DINO}$\uparrow$ & \textbf{Face-S}$\uparrow$ & \textbf{mIoU}$\uparrow$ & \textbf{EPE}$\downarrow$ & \textbf{CLIP-T}$\uparrow$ \\ 
        \midrule
        \multicolumn{7}{c}{\textit{\textbf{Single-Subject Mode}}} \\
        \midrule
        0.01 & \underline{0.737} & \textbf{0.505} & \underline{0.279} & \textbf{0.560} & 9.85 & \underline{0.307} \\
        \textbf{0.10 (Ours)} & \textbf{0.739} & \underline{0.499} & \textbf{0.301} & \underline{0.558} & \textbf{9.31} & \textbf{0.308} \\ 
        0.25 & 0.735 & 0.492 & 0.272 & 0.555 & \underline{9.65} & \underline{0.307} \\
        0.50 & 0.718 & 0.482 & 0.223 & 0.541 & 9.75 & 0.306 \\
        1.00 & 0.674 & 0.350 & 0.120 & 0.350 & 25.00 & 0.280 \\
        \midrule
        \multicolumn{7}{c}{\textit{\textbf{Multi-Subject Mode}}} \\
        \midrule
        0.01 & \underline{0.718} & \underline{0.518} & 0.322 & \underline{0.557} & 6.70 & \textbf{0.306} \\
        \textbf{0.10 (Ours)} & \textbf{0.720} & \textbf{0.524} & \underline{0.329} & \textbf{0.570} & \underline{6.08} & \textbf{0.306} \\ 
        0.25 & 0.714 & 0.515 & \textbf{0.331} & 0.538 & \textbf{5.95} & \textbf{0.306} \\
        0.50 & 0.710 & 0.504 & 0.295 & 0.530 & 6.73 & \underline{0.305} \\
        1.00 & 0.692 & 0.380 & 0.287 & 0.380 & 15.06 & 0.280 \\
        \bottomrule
    \end{tabular}
    }
\end{table}

\noindent\textbf{Design choices for latent identity reward model.}\quad
We investigate the impact of different design choices on the pairwise classification accuracy of the latent identity reward model in Table~\ref{tab:ablation_reward_model}.
The accuracy is measured on the test set of win-lose pairs, where a prediction is correct if the reward score of the winner video is higher than that of the loser.
\textit{(i)} Regarding the optimization objective, our default Binary Cross-Entropy (BCE) loss yields superior performance (Avg 0.720) compared to the Bradley-Terry (BT) model~\cite{bradley1952rank} (Avg 0.656). Notably, the BT loss exhibits significant instability at early timesteps ($t \in [0, 0.2]$).
\textit{(ii)} For the image injection strategy, employing the reference image as the Query is critical. Treating it as Key/Value (KV) results in a catastrophic accuracy drop (to 0.455), indicating that the reference image must actively attend to the noisy latents to effectively discern identity features.
\textit{(iii)} Regarding the parameter tuning scope, freezing the text and patch embeddings outperforms fine-tuning them, suggesting that preserving pre-trained priors prevents overfitting and is sufficient for effective reward modeling.

\noindent\textbf{Effect of the range of timestep $t_m$ in LIReFL.}\quad
Table~\ref{tab:ablation_refl_reward_range} investigates the influence of the range of timestep $t_m$ on model performance.
We compare sparse feedback on the last 3 steps against dense feedback across all timesteps.
In single-subject mode, this full-range strategy yields comprehensive improvements across all metrics.
In multi-subject scenarios, it further enhances identity metrics while maintaining precise motion control.
These findings indicate that providing reward feedback at arbitrary timesteps is essential to fully leverage the potential of LIReFL, enhancing identity fidelity throughout the generation process.

\noindent\textbf{Effect of loss weight $\lambda_2$ in LIReFL.}\quad
Table~\ref{tab:ablation_refl_weight} illustrates the impact of varying the feedback learning strength $\lambda_2$.
We observe that values within the range of $[0.01, 0.25]$ yield robust overall performance.
Notably, identity fidelity metrics, such as R-DINO and Face-S, consistently surpass those of the SFT-only model (see Table~\ref{tab:ablation_sft_components}), indicating that identity reward feedback effectively enhances customization and refines appearance details.
However, performance begins to decline when $\lambda_2$ increases to 0.50. Furthermore, at $\lambda_2 = 1.00$, the model suffers from reward hacking, where it finds shortcuts to maximize the reward at the expense of visual realism and motion coherence.
This suggests that excessively strong feedback disrupts the generative process, a phenomenon also observed in previous reinforcement learning studies~\cite{skalse2022defining, mi2025video}.
Balancing robust identity preservation with motion control, we adopt $\lambda_2 = 0.10$ as the optimal configuration for our final model.

\section{Conclusion}
\label{sec:conclusion}
In this work, we present \framework, a unified framework that achieves harmonious multi-subject customization with omni-motion control, encompassing global and local object motion as well as camera movement.
By introducing a progressive two-stage training paradigm, we effectively resolve the conflict between identity preservation and complex motion control.
Specifically, to ensure robust and precise controllability, we incorporate a condition-aware 3D RoPE to coordinate heterogeneous inputs and a hierarchical motion injection strategy to enhance global motion guidance.
To further eliminate control ambiguity in multi-subject scenarios, we propose group and role embeddings to explicitly bind motion signals to their corresponding identities.
Furthermore, we design a latent identity reward feedback learning paradigm, leveraging a VDM-based latent identity reward model to prioritize motion-aware identity preservation aligned with human preferences.
To advance the field, we design a comprehensive automated data construction pipeline and establish the \textbf{DreamOmni Bench}, a holistic benchmark for evaluating multi-subject and omni-motion control.
Extensive experiments demonstrate that \frameworkplain significantly outperforms state-of-the-art methods in generating high-quality, customizable videos with precise and flexible motion control.

\bibliographystyle{IEEEtran}
\bibliography{main}

\end{document}